\documentclass[journal,10pt]{IEEEtran}
\usepackage{cite}

\ifCLASSINFOpdf
   \usepackage[pdftex]{graphicx}
\else
   \usepackage[dvips]{graphicx}
\fi
\ifCLASSOPTIONcompsoc
  \usepackage[caption=false,font=normalsize,labelfont=sf,textfont=sf]{subfig}
\else
  \usepackage[caption=false,font=footnotesize]{subfig}
\fi
\usepackage{url}

\usepackage{floatrow}
\usepackage{multirow}
\usepackage[tableposition=top]{caption}

\newfloatcommand{capbtabbox}{table}[][\FBwidth]
\newsavebox\CBox
\def\textBF#1{\sbox\CBox{#1}\resizebox{\wd\CBox}{\ht\CBox}{\textbf{#1}}}
\newcommand{\etal}{\textit{et al}.}
\newcommand{\ie}{\textit{i}.\textit{e}.}
\newcommand{\eg}{\textit{e}.\textit{g}.}

\begin{document}
%
\title{Pancreas Segmentation in CT and MRI Images via Domain Specific Network Designing and Recurrent Neural
Contextual Learning}
%
%
%

\author{Jinzheng~Cai, Le~Lu, Fuyong~Xing, and~Lin~Yang
\thanks{J. Cai, F. Xing, and L. Yang are with the J. Crayton Pruitt Family Department of Biomedical Engineering, University of Florida, Gainesville, FL 32611 USA (e-mail: jimmycai@ufl.edu; shampool@ufl.edu; f.xing@ufl.edu; lin.yang@bme.ufl.edu).}
\thanks{L. Lu is with Nvidia Corporation, 2788 San Tomas Expy, Santa Clara, CA 95051, USA (e-mail: lel@nvidia.com).}}

\maketitle

\begin{abstract}
Automatic pancreas segmentation in radiology images, \eg{}, computed tomography (CT) and magnetic resonance imaging (MRI), is frequently required by computer-aided screening, diagnosis, and quantitative assessment. Yet pancreas is a challenging abdominal organ to segment due to the high inter-patient anatomical variability in both shape and volume metrics. Recently, convolutional neural networks (CNNs) have demonstrated promising performance on accurate segmentation of pancreas. However, the CNN-based method often suffers from segmentation discontinuity for reasons such as noisy image quality and blurry pancreatic boundary. From this point, we propose to introduce recurrent neural networks (RNNs) to address the problem of spatial non-smoothness of inter-slice pancreas segmentation across adjacent image slices. To inference initial segmentation, we first train a 2D CNN sub-network, where we modify its network architecture with deep-supervision and multi-scale feature map aggregation so that it can be trained from scratch with small-sized training data and presents superior performance than transferred models. Thereafter, the successive CNN outputs are processed by another RNN sub-network, which refines the consistency of segmented shapes. More specifically, the RNN sub-network consists convolutional long short-term memory (CLSTM) units in both top-down and bottom-up directions, which regularizes the segmentation of an image by integrating predictions of its neighboring slices. We train the stacked CNN-RNN model end-to-end and perform quantitative evaluations on both CT and MRI images.
\end{abstract}


%
\IEEEpeerreviewmaketitle

\section{Introduction}
%
%
%
%
\IEEEPARstart{D}{etecting} unusual volume changes and monitoring abnormal growths in pancreas using medical images is a critical yet challenging diagnosis task. This would require to delineate pancreas from its surrounding tissues in radiology images, \eg{}, computed tomography (CT) and magnetic resonance imaging (MRI) scans. Having pancreas accurately segmented from 3D scans delivers more reliable and quantitative representations than simple cross-section diameter measurements, which may produce the precision segmentation based biomarkers, such as volumetric measurements and 3D shape/surface signatures. Moreover, automated rapid and accurate segmentation of pancreas on the scale of processing thousands of image scans can facilitates new protocols, findings, and insights for clinical trials. On the other hand, manual pancreas segmentation is very expensive and sometimes even intractable on the dataset at a very large scale. In this paper, to fulfill this practical and important demand, we improve the existing convolutional neural networks (CNN) based segmentation work with novel methodologies, i.e., new CNN module architecture, convolutional-recurrent neural contextual regularization and a new direct segmentation loss, to achieve significantly boosted performances in both CT and MRI imaging modalities. 

\begin{figure}[t!]
	\begin{center}
		\includegraphics[width=.9\linewidth, page=6, trim={0cm 9.5cm 17cm 0cm}, clip]{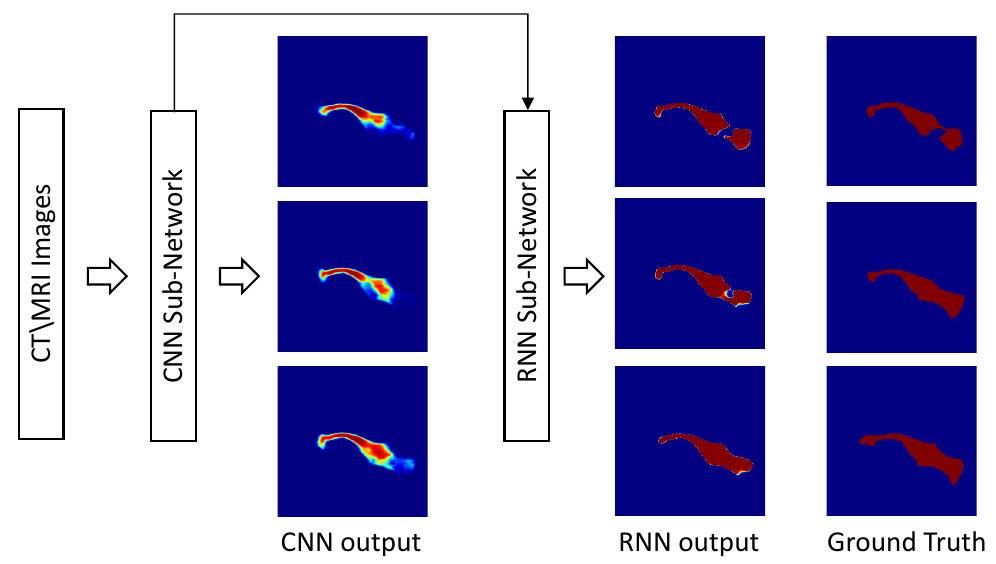}
	\end{center}
	\caption{Overview of the proposed model architecture. CT/MRI image sequence is processed by the CNN-RNN segmentation model. First, the CNN sub-network inferes probability maps of the 2D input slices. The RNN sub-network then refines the segmentations to improve the intra-slice shape continuity. We note that the CNN-RNN atacked model is trained end-to-end.}
	\label{fig:overview}
\end{figure}

~~~One major group of related work on automatic pancreas segmentation in CT images is based on top-down multi-atlas registration and label fusion (MALF)~\cite{Oda2016,Tong201592,wolz2013automated,Karasawa2017Multi}. Due to the high deformable shape and vague boundaries of the pancreas in CT scans from various patients, their reported segmentation accuracy results (measured in Dice Similarity Coefficient or DSC) range from 69.6$\pm$16.7\% \cite{wolz2013automated} to 78.5$\pm$14.0\%~\cite{Karasawa2017Multi,Oda2016} under leave-one-patient-out (LOO) evaluation protocol. On the other hand, bottom-up deep CNN based pancreas segmentation work~\cite{Cai2016,FaragLRLTS15,roth2015deeporgan,RothLFSS16,ZhouXSFY16,Roth2017} have revealed promising results and steady performance improvements, e.g., from 71.8$\pm$10.7\%~\cite{roth2015deeporgan}, 78.0$\pm$8.2\%~\cite{RothLFSS16}, to 81.3$\pm$6.3\%~\cite{Roth2017} evaluated using the same NIH 82-patient CT dataset under 4 fold cross-validation (CV).

~~~In comparison, deep CNN approaches appear to demonstrate noticeably higher segmentation accuracies and numerically more stable results (significantly lower in standard deviation, or std) than their MALF counterparts. \cite{RothLFSS16,Roth2017} are built upon the fully convolutional neural network (FCN) architecture~\cite{long2014fully} and its variant~\cite{xie2015holistically}. However,~\cite{RothLFSS16,Roth2017} both rely on post-processing with random forest to further refine CNN's outputs, which can not propagate errors back to the CNN model. Similarly, for pancreas segmentation on a 79-patient MRI dataset,~\cite{Cai2016} achieves 76.1$\pm$8.7\% in DSC, where graph-based result fusion is applied. Therefore an end-to-end trainable deep learning model for pancreas segmentation may be more desirable to achieve superior results. Additionally, deep CNN based bottom-up pancreas segmentation methods also have significant advantages on run-time computational efficiency, such as $2\sim4$ hours in~\cite{Karasawa2017Multi} versus $2\sim3$ minutes for~\cite{Roth2017} to process a new segmentation case.

~~~From the preliminary version of this paper~\cite{Cai2017Improving}, we propose a compact-sized CNN architecture with recurrent neural contextual learning for improved pancreas segmentation. In this paper, the segmentation performance of the 2D CNN sub-network is further improved by combining deeply-supervised model training~\cite{DBLP:conf/aistats/LeeXGZT15} with dedicated multi-scale aggregation~\cite{yu_2015_msa} of the low-level features, partially inspired by the U-Net architecture in~\cite{Ronneberger2015}. Our implementation implies that a much smaller network that is trained from scratch on the target dataset~\cite{roth2015deeporgan}, \eg{}, pancreas CT/MRI scans, outperforms the full-sized UNet~\cite{Ronneberger2015} model, fine-tuned from the other image domain. Subsequently, the sub-network for modeling pancreatic shape continuity is strengthened by taking slices from both the top-down and bottom-up axial directions. This is achieved by stacking two layers of convolutional long short-term memory (CLSTM)~\cite{ShiCWYWW15}, which encode neighboring contextual constraints in two opposite directions and we name this structure BiRNN. A similar bi-directional CLSTM architecture is also proposed in~\cite{ChenYZAC16}; however, we apply BiRNN explicitly at the output of 2D CNN sub-network to regularize the shape consistency of segmentation results. More importantly, the implementation of BiRNN is computationally efficient because it introduces only a small amount of model parameters (72 variables per CLSTM layer) and requires convolution operations on only one channel of the feature map. On such simplified problem design, the light-weighted BiRNN performs efficient training convergence. Note that another major reason why we design stacked 2D CNN-RNN architecture to process organ segmentation is that the inter-slice thickness in low-dose CT scans (of screening or repetitive oncology imaging protocols) and MRI volumes range from 2.5 to 5 mm. The out-of-plane spatial smoothness in the axial direction is more sparse than in-plane resolutions (e.g., $\leq$1 mm) which would significantly limit the effectiveness of 3D CNN models.

~~~Next, we present a novel segmentation-direct loss function to train our end-to-end deep neural network models by minimizing the Jaccard index (JI) between any pair of the ground truth annotated pancreas mask and its corresponding segmentation output mask. The standard practice in FCN image segmentation models~\cite{long2014fully,xie2015holistically,Cai2016,RothLFSS16} use a loss function to sum up the cross-entropy loss at each voxel or pixel. A segmentation-direct loss function can avoid the data balancing issue during CNN training between the pixel counts residing in positive (pancreas) or negative (background) regions. The pancreas region normally only occupies a very small fraction of each CT/MRI slice. Furthermore, there is no need to calibrate the optimal probability threshold to achieve the best possible binary pancreas segmentation results from the FCN's probabilistic outputs~\cite{long2014fully,xie2015holistically,Cai2016,RothLFSS16}. Similar segmentation metric based loss functions using Dice similarity coefficient (DSC) are concurrently proposed and investigated in~\cite{MilletariNA16,ZhouXSFY16}.

~~~We have extensively and quantitatively evaluated our proposed CNN-RNN pancreas segmentation model and its ablated variants using both a CT (82 patients) and an MRI (79 patients) dataset, under 4-fold cross-validation (CV). Our complete model outperforms in DSC by noticeable margins, comparing to previous state-of-the-arts~\cite{Cai2016,RothLFSS16,Roth2017,ZhouXSFY16}. Although our compact CNN model design and shape continuity/regularization learning model is only tested on pancreas segmentation, the approach is directly generalizable to other organ segmentation tasks using medical imaging. Code and models for our experiments will be released upon publication.

\section{Related Work}
\subsection{Segmentation Model for Object with Complex Shape}

\begin{figure}[t!]
	\centering
	\subfloat[{HNN}]{%
		\includegraphics[width=.24\linewidth]{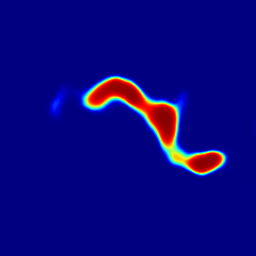}
	}
	\subfloat[{UNet}]{%
		\includegraphics[width=.24\linewidth]{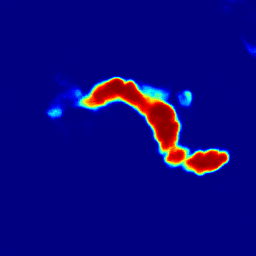}
	}
	\subfloat[{PNet-MSA}]{%
		\includegraphics[width=.24\linewidth]{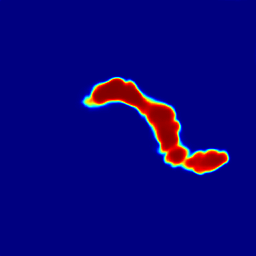}
	}
	\subfloat[ground truth]{%
		\includegraphics[width=.24\linewidth]{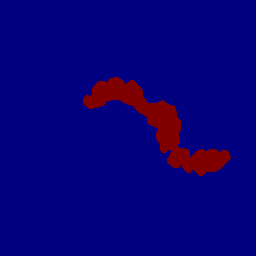}
	}
	\caption{Comparison of output probability maps between {UNet}, {HNN}, and the proposed {PNet-MSA}, where {PNet-MSA} presents more boundary details and less false positive detections comparing to {HNN} and {UNet}, respectively.}
	\label{fig:examplar-case}
\end{figure}
With the goal of obtaining the detailed and accurate segmentation outcome for complex and difficult objects, a number of deep learning approaches have been proposed recently~\cite{long2014fully,chen2014semantic,zheng2015conditional,xie2015holistically,Ronneberger2015}. Some of these models regularize deep learning output with the original image appearance that the image pixels sharing similar colors (or intensity values) probably come from the same object category. Conditional random field or CRF~\cite{chen2014semantic,zheng2015conditional} is proposed to handle this scenario. Other methods propose to learn more localized deep features. For instance, deep supervision is proposed in~\cite{xie2015holistically,DBLP:conf/aistats/LeeXGZT15}, forcing all convolutional layers to learn effective and consistent low-level representations to capture local image features, \eg{}, edge and object boundary. Meanwhile, UNet architecture in~\cite{Ronneberger2015} is presented to make full use of low-level convolutional feature maps by projecting them back to the original image size which delineates object boundaries with details. To combine virtues of both network design, we propose to adopt and build upon the previously proposed UNet~\cite{Ronneberger2015} and holistically nested network (HNN)~\cite{xie2015holistically} in this paper due to their end-to-end trainable nature as well as excellent segmentation performances. As shown in Fig.~\ref{fig:examplar-case}, the output of UNet catches many details implying that the dedicated backward propagation combining convolutional layer features of multiple scales is helpful to accurate pancreas segmentation. However, the probability map of a well-trained UNet can still be blurring, along with false positive detections. 

\subsection{Segment Pancreas in 3D}
Previous work~\cite{Cai2016,RothLFSS16,ZhouXSFY16} perform deep 2D CNN segmentation on either CT or MRI image or slice independently. Organ segmentation in 3D CT and MRI scans can also be performed by directly taking cropped 3D sub-volumes as input~\cite{MilletariNA16,MerkowKMT16,KamnitsasLNSKMR16}. Even at the expense of being computationally expensive and prone-to-overfitting~\cite{ZhouXSFY16}, the result of very high segmentation accuracy has not been reported for complexly shaped organs~\cite{MerkowKMT16}, or small anatomical structures~\cite{KamnitsasLNSKMR16}. Despite more demanding memory requirement, 3D CNN approaches deserve more investigation for future work. \cite{ChenYZAC16,StollengaBLS15} use hybrid CNN-RNN architectures to process/segment sliced CT or MRI images in sequence. There are no spatial shape continuity or regularization constraints enforced among successive slices. 

~~~In this work, we model 3D pancreas segmentation by first following the framework by training 2D slice-based CNN models for pancreas segmentation. Once this step of CNN training converges, inspired by the sequence modeling for precipitation nowcasting in~\cite{ShiCWYWW15}, a convolutional long short-term memory (CLSTM) network is further added to the output layer of the deep CNN to explicitly capture and constrain the contextual segmentation smoothness across neighboring image slices. We may argue that the CLSTM based sequence modeling captures more complex contextual segmentation information than 3D convolutional kernels. It is empirically observed that the improved segmentation performance by learning convolution among different CT slices using multi-channel CNN saturates with three successive frames or images~\cite{ZhouXSFY16}. Then the whole integrated network can be end-to-end fine-tuned via stochastic gradient descent (SGD) until convergence. The CLSTM module will modify the segmentation results produced formerly by CNN alone, by taking the initial CNN segmentation results of successive axial slices (in either superior or interior direction) into account. The final segmented pancreas shape is constrained to be consistent among adjacent slices, as a good trade-off between 2D and 3D deep models. This sequence learning representation demonstrates the analogy in video content/activity comprehension, from using direct 3D convolutional kernels \cite{Karpathy2014Large} to capture atomic actionlets towards the more  ``flexible CNN-encoding followed by RNN-decoding'' paradigm \cite{Ng2015Beyond} where long-range sequential dependencies can be modeled.

\section{Method} \label{section:method}

\subsection{CNN Sub-Network}
Delineating the pancreas boundary from its surrounding structures in CT/MRI imaging can be challenging due to its complex visual appearance and ambiguous outlines. For example, Fig.~\ref{fig:examplar-case} displays a cross-section area of a pancreas in {CT}, where the pancreas shares similar intensity values with other soft tissues and its boundary is blurry where touching with abdominal organs. Strong existence of visual ambiguity for pancreas segmentation in {CT} (and {MRI}) imaging modalities demonstrate very different image statistics from the problem of semantic object segmentation in natural images. Direct transfer learning of natural image pre-trained CNNs to medical domain might be suboptimal. More specifically, when fine-tune an ImageNet pre-trained model~\cite{simonyan2014very} for pancreas segmentation, we observe that the training loss drops fast, indicating that the top CNN model layers can capture the hierarchical (conceptual) appearance representation of pancreas sufficiently. However, the magnitudes of gradients (back-propagated from the training error) decrease rapidly to leave the bottom CNN layers not well tuned. This situation becomes even worse when the gradient-vanishing problem occurs during model training. To alleviate this undesirable effect, we propose and exploit a new method to enable CNN that can be effectively trained from scratch, even with small-sized medical datasets.

~~~To make bottom CNN layers be more discriminative to pancreatic and non-pancreatic pixels,  we train the {CNN} model from scratch. Toward this end, in our preliminary work~\cite{Cai2017Improving}, we design a {CNN} network architecture with much smaller model size ($\sim10$\% of parameters) than the models in~\cite{xie2015holistically,long2014fully} so as to stop the network form over-fitting to the small-sized training set and accelerate its convergence. Meanwhile, deep-supervision~\cite{DBLP:conf/aistats/LeeXGZT15} is used introducing strong training supervisions at the bottom {CNN} layers. In preliminary, such network design performs superior to {HNN}~\cite{xie2015holistically} and {UNet}~\cite{Ronneberger2015} that transferred from other tasks. To further extend our original network design, in this work, we investigate to upgrade the CNN model with multi-scale feature map aggregation (MSA), which is largely inspired by the network architecture of UNet~\cite{Ronneberger2015}. Since we originally present our network design for pancreas segmentation, we refer to it as PNet and its extension proposed in this work as {PNet-MSA}.

\begin{figure*}[t!]
	\centering
	\includegraphics[width=.8\linewidth, page=2, trim={0cm 6.2cm 7cm 0cm}, clip]{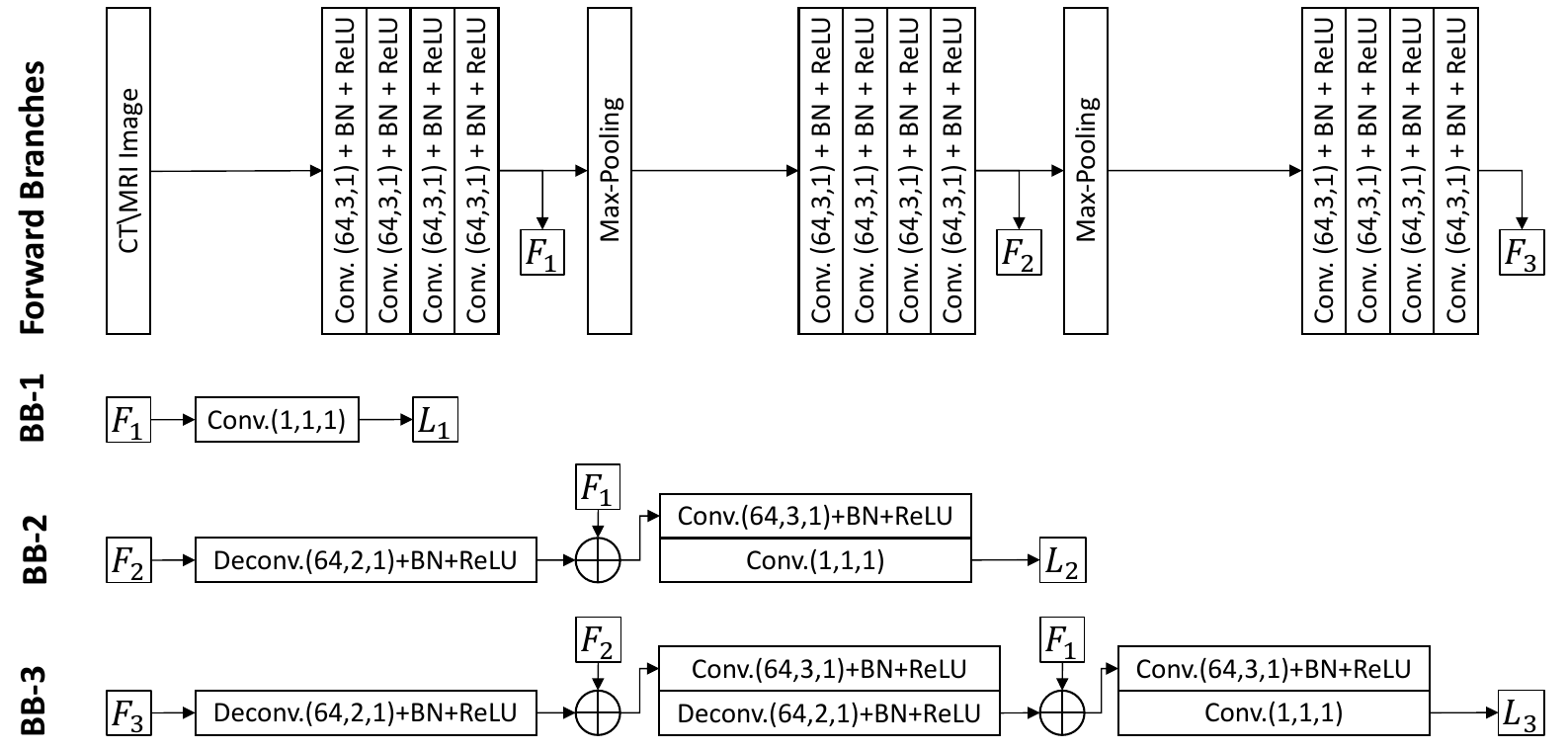}
	\caption{The main construction units of the proposed PNet-MSA.  In practice, we use 5 unit modules for pancreas segmentation and the first 3 are visually depicted here. The \emph{Deconv., Conv., BN, ReLU} and $\bigoplus$ present the CNN model structures as the deconvolution, convolution, batch normalization, rectified linear unit, and concatenation layers, respectively. Hyperparameters for \emph{Deconv.} and \emph{Conv.} are given in the bracket as the number of output channels, kernel size, and kernel stride. The $1^{st}$, $2^{nd}$, and $3^{rd}$ forward branches output feature maps $F_1$, $F_2$, and $F_3$ in respective. The ``BB'' is short for backward branch, and $L_1$, $L_2$, and $L_3$ are the corresponding segmentation losses defined in Eq.~\ref{eq:loss}.}
	\label{fig:pnet-architecture}
\end{figure*}

~~~To train the {PNet-MSA} from scratch, we design the model as a stack of units with the same network architecture. For each unit, it is first initialized with MSRA initialization~\cite{DBLP:conf/iccv/HeZRS15} and then added as top layers of the network for end-to-end training. Specifically, each unit module consists a forward branch and a corresponding backward branch. The forward branch contains 4 layers of (convolution+batch normalization+ReLU) combinations. The backward branch first performs MSA by concatenating the current forward branch output with former outputs via deconvolution, and then the concatenated MSA features are fused to generate the segmentation result. Fig.~\ref{fig:pnet-architecture} visually depicts 3 stacked unit modules, where between each pair of the unit modules, a pooling layer, \eg{}, max-pooling, is inserted into the forward branches to increase the receptive field size of the top network layers. 

~~~To formally present our network design, we first denote the $k$-th unit module a nested mapping function, 
\begin{equation}
\label{eq:unit-module}
U_k = B_k(F_1(X_1; \theta_1^f),\ldots,F_k(X_k; \theta_k^f); \theta_k^b ),
\end{equation}
where $F_k(\cdot;\theta_k^f)$ and $B_k(\cdot;\theta_k^b)$ represents the $k$-th forward and backward branch, and $\theta_k^f$, $\theta_k^b$ are the corresponding model parameters that can be optimized via training. The $X_k \in R^{d^1_k\times d^2_k\times d^3_k}$ is the input, which is either output of the $(k-1)$-th forward branch or the input {CT/MRI} image (\ie{}, $X_1$), where $d^1_k$, $d^2_k$, and $d^3_k$ represent the correponding tensor dimensions. And $U_k$ it the output probability map for pancreas segmentation. On $U_k$, a segmentation loss is measured by,
\begin{equation}
\label{eq:loss}
\ell_k = \mathcal{L}(U_k, Y), 
\end{equation}
where $Y$ is the ground truth for model training, and we will define the loss function $\mathcal{L}(\cdot, Y)$ in Sec.~\ref{sec:jacloss}. Thus, each unit module has its own update gradient flow starting from its loss branch $L_k$. This design is important because it introduces deep-supervision to the bottom CNN layers, and it enable us to train the PNet-MSA layer by layer from swallow to deep. More specifically, PNet-MSA training starts from $k=1$, and $k$ is increased by $1$ when $L_k$-plot converges. The ultimate segmentation results will be a weighted combination of the unit module outputs,
\begin{equation}
\hat{Y} = \sum_{k=1}^{K} w_k U_k,
\end{equation}
and the overall objective for PNet-MSA is, 
\begin{equation}
\ell = \mathcal{L}(\hat{Y},Y) + \sum_{k=1}^{K} \ell_k, 
\end{equation}
where we find $K=5$ works the best for pancreas segmentation.

~~~Although PNet-MSA may be extended from processing the 2D input image to 3D~\cite{cciccek20163d}, we maintain its 2D architecture as training and inference since the 3D versions can be computationally expensive, with no significant improvement in performance~\cite{ZhouXSFY16}. However, our 2D model takes 3-connected-slices as its input when given the segmentation ground truth mask of the middle slice. As explained in Sec.~\ref{sec:RNN}, our PNet-MSA is transformed into a light-weighted 3D model with RNN stacked to the end of it, which allows our model to capture 3D imaging information with minor extra computational loads. This is in a similar spirit to employ RNN to regulate, process and aggregate CNN activations for video classification and understanding \cite{Ng2015Beyond}.

\subsection{Recurrent Neural Contextual Learning} \label{sec:RNN}
\begin{figure}[!t]
	\centering
	\subfloat[$\hat{Y}_{\tau-2}$]{%
		\includegraphics[width=.19\linewidth]{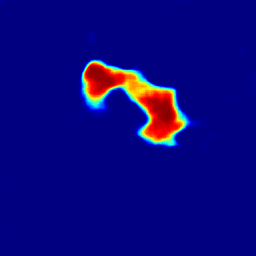}
	}
	\subfloat[$\hat{Y}_{\tau-1}$]{%
		\includegraphics[width=.19\linewidth]{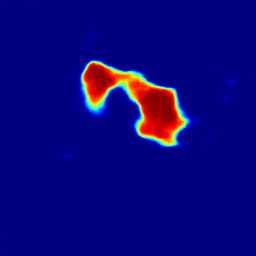}
	}
	\subfloat[$\hat{Y}_{\tau}$]{%
		\includegraphics[width=.19\linewidth]{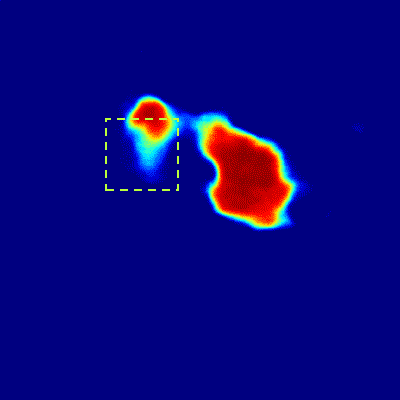}
	}
	\subfloat[$\hat{Y}_{\tau+1}$]{%
		\includegraphics[width=.19\linewidth]{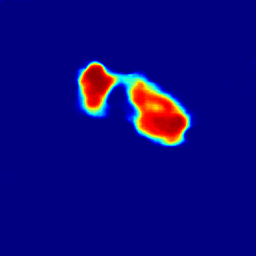}
	}
	\subfloat[$\hat{Y}_{\tau+2}$]{%
		\includegraphics[width=.19\linewidth]{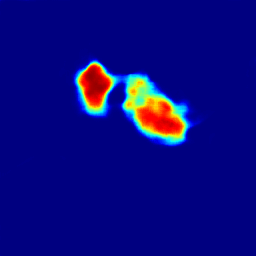}
	}
	\\
	\subfloat[CLSTM]{%
		\includegraphics[width=0.49\linewidth, height=4.5cm, page=3, trim={0cm 11.5cm 26.5cm 0cm}, clip]{figures/Figures.pdf}
	}
	\subfloat[Bi-direction CLSTM]{%
		\includegraphics[width=0.49\linewidth, height=4.5cm, page=4, trim={0cm 10.9cm 24cm 0cm}, clip]{figures/Figures.pdf}
	}
	\\
	\subfloat[$\bar{Y}_{\tau-2}$]{%
		\includegraphics[width=.19\linewidth]{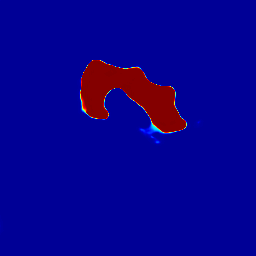}
	}
	\subfloat[$\bar{Y}_{\tau-1}$]{%
		\includegraphics[width=.19\linewidth]{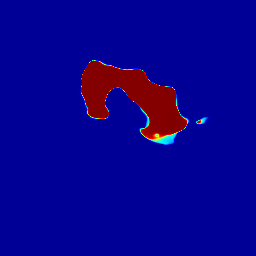}
	}
	\subfloat[$\bar{Y}_{\tau}$]{%
		\includegraphics[width=.19\linewidth]{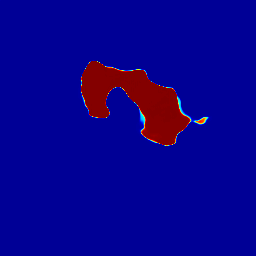}
	}
	\subfloat[$\bar{Y}_{\tau+1}$]{%
		\includegraphics[width=.19\linewidth]{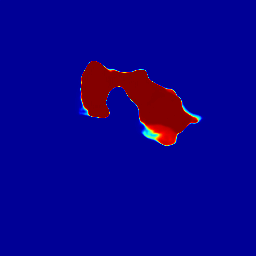}
	}
	\subfloat[$\bar{Y}_{\tau+2}$]{%
		\includegraphics[width=.19\linewidth]{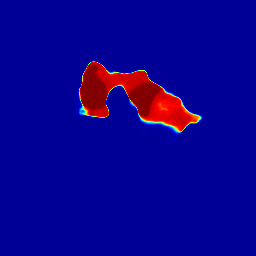}
	}
	\caption{The main construction units of the proposed RNN sub-network and its input/output segmentation sequence. The sequence of CNN sub-network outputs is shown in the first row (a-e), is taken as the input of the bi-direction CLSTM (g), which is an RNN architecture composed of 2 layers of CLSTM (f) working in opposite directions. The third row (h-l) presents the corresponding output sequence, which is sharp and clean. Note that the missing pancreatic part in $\hat{Y}_\tau$ (c), in the green dashed box, is recovered by shape continuity modeling in $\bar{Y}_\tau$ (j). For visual clearity, we ommit the input $\hat{Y}_{(\cdot)}$ in the bi-direction CLSTM (g), which is same as in (f).}
	\label{fig:contextual-learning}
\end{figure}

From above, the CNN sub-network processes pancreas segmentation on individual 2D image slices, delivering remarkable performance on the tested CT and MRI datasets. However, as shown in the first row of Fig.~\ref{fig:contextual-learning}, the transition among the resulted CNN pancreas segmentation regions in consecutive slices may not be smooth, often implying that segmentation failure occurs. Adjacent CT/MRI slices are expected to be correlated to each other thus segmentation results from successive slices need to be constrained for shape continuity and regularization. The model for 3D object segmentation is required to be able to detect and recover abnormally lost part inside slices (see $\hat{Y}_{\tau}$ in Fig~.\ref{fig:contextual-learning}).

~~~To achieve this, we concatenate an RNN sub-network, which is originally designed for sequential data processing, to the CNN sub-network for modeling inter-slice shape continuity and regularization. That is, we slice any 3D CT (or MRI) volume into an ordered sequence of 2D images and process to learn the segmentation shape continuity among neighboring image slices with a typical RNN architecture, the long short-term memory (LSTM) unit. However, the standard LSTM requires vectorized input which would sacrifice the spatial information encoded in the output of CNN. We, therefore, utilize the convolutional LSTM (CLSTM) model~\cite{ShiCWYWW15} to preserve the 2D image segmentation layout by CNN. As shown in Fig.~\ref{fig:contextual-learning}, $H_{\tau}$ and $C_{\tau}$ are the hidden state and cell output of CLSTM in respective at the $\tau$-th slice. The current cell output $C_\tau$ is computed based on both of the former cell hidden state $H_{\tau-1}$ and the current CNN output $\hat{Y}_{\tau}$. Then, $H_{\tau}$ will be calculated from $C_{\tau}$ and used to produce the next cell output $C_{\tau+1}$. Contextual information is propagated from slice $\tau$ to $\tau+1$ through convolutional operations. 

~~~Our Convolutional LSTM based inter-slice or sequential segmentation shape continuity learning is intuitive. Segmentation results of the former image slices are encoded in the cell hidden state $H_{\tau-1}$. Values of $C_\tau$ is decided by taking $H_{\tau-1}$ and $\hat{Y}_{\tau}$ together into consideration. If position $p_i$ in $\hat{Y}_\tau$ is predicted as pancreatic tissue by the CNN sub-network, and the same position in $H_{\tau-1}$ are also encoded as pancreatic tissue, then with high confidence that position $p_i$ should be a pancreatic pixel in $C_{\tau}$, and vice versa. As a result, CLSTM not only recovers missing segmentation parts but also outputs more confident probability maps than the original CNN sub-network. Formally, the CLSTM unit is formulated, 
\setlength{\arraycolsep}{0.0em}
\begin{eqnarray}
I_\tau& {}={} &\sigma(W_{yi}{*}\hat{Y}_\tau + W_{hi}{*}H_{\tau-1} + W_{ci}{\odot}C_{\tau-1} + b_i), \\
F_\tau& {}={} &\sigma(W_{yf}{*}\hat{Y}_\tau + W_{hf}{*}H_{\tau-1} + W_{cf}{\odot}C_{\tau-1} + b_f), \\
C_\tau& {}={} &f_\tau{\odot}C_{\tau-1}{+}i_\tau{\odot}\tanh(W_{yc}{*}\hat{Y}_\tau{+}W_{hc}{*}H_{\tau-1}{+}b_c), \\
O_\tau& {}={} &\sigma(W_{yo}{*}\hat{Y}_\tau + W_{ho}{*}H_{\tau-1} + W_{co}{\odot}C_\tau + b_o), \\
H_\tau& {}={} &o_\tau{\odot}\tanh(c_\tau),
\end{eqnarray} 
where $*$ represents convolution operation, and $\odot$ denotes the Hadamard product. Gates $I_\tau$, $F_\tau$, $O_\tau$ are the input, forget, and output following the original definition of CLSTM. $W_{(\cdot)}, b_{(\cdot)}$ are weights and bias in the corresponding CLSTM unit that need model optimization. Finally, $\sigma(\cdot)$ and $\tanh(\cdot)$ denote the sigmoid and hyperbolic tangent activation function, respectively. 

{\bf Bi-directional Contextual Regularization:} We then propose to use a bi-direction extension of the CLSTM. For pancreas as well as other organs, its shape in the current slice is constrained by slices from not only the former ones but also its followings. The contextual information to input could be doubled if the shape regularization is taken in both directions leading to a further improvement. Two layers of CLSTM are stacked working in two opposite directions as shown in Fig.~\ref{fig:contextual-learning}. Then, outputs of the two layers, one in $\tau^-$-direction and the other in $\tau^+$-direction, is weighted combined as final segmentation output, 
\begin{equation}
\bar{Y}_{\tau} = \sum_{i \in \{-,+\} }\lambda^{i} O^{i}_{\tau},
\end{equation}
where $i$ represents the $\tau^-$ and $\tau^+$ directions, and $\lambda^{i}$ is the learned weights when combining CLSTM outputs from both directions. Thus, the bi-direction design of shape continuity modeling permits to explicitly enforce the pancreas segmentation spatial smoothness and higher-order inter-slice regularization.

\subsection{Jaccard Loss} \label{sec:jacloss}

\begin{figure*}
\begin{floatrow}
\ffigbox[.8\linewidth]{%
	\includegraphics[width=.85\linewidth, height=.63\linewidth, page=5, trim={0cm 7.3cm 21cm 0cm}, clip]{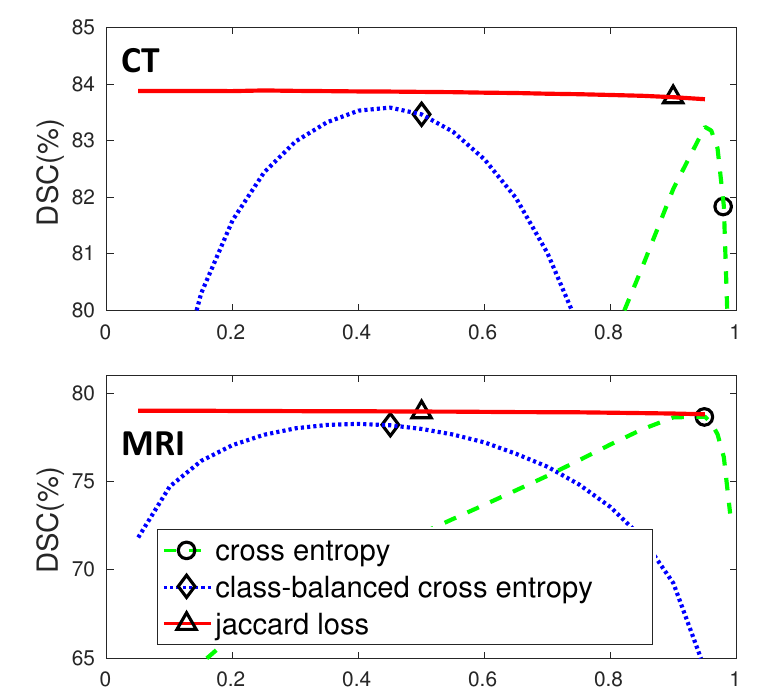}
}{%
	\caption{Plot of the threshold v.s. DSC: the proposed jaccard loss (JAC) performs the steadiest across thresholds in the range of $[0.05,0.95]$ comparing to the cross entropy (CE) and the class-balanced cross entropy (CBCE).}%
	\label{fig:losscurves}
}
\capbtabbox{%
	\begin{tabular}{ccc} \hline
		\multirow{2}{*}{\bf Loss} & \multicolumn{2}{c}{mean$\pm$stdv.[min,max]} \\ \cline{2-3}
		& DSC(\%) & JI(\%)\\ \hline
		\multicolumn{3}{c}{\bf CT} \\ \hline
		CE       & 83.5$\pm$5.6 [59.3,91.1] & 72.0$\pm$7.70 [42.2,83.6] \\
		CBCE & 83.2$\pm$5.7 [57.2,90.3] & 71.6$\pm$7.80 [40.1,82.4] \\
		JAC  & \textBF{83.7$\pm$5.4 [58.4,90.4]} & \textBF{72.3$\pm$7.50 [41.3,82.4]} \\ \hline
		\multicolumn{3}{c}{\bf MRI} \\ \hline
		CE	   & 80.0$\pm$7.60 [50.7,89.9] & 67.3$\pm$9.80 [34.0,81.6] \\
		CBCE     & \textBF{80.2$\pm$7.20 [53.6,90.5]} & \textBF{67.6$\pm$9.50 [36.6,82.7]} \\
		JAC  & \textBF{80.2$\pm$7.90 [51.2,90.1]} & \textBF{67.6$\pm$10.3 [34.4,82.0]} \\ \hline
	\end{tabular}
}{%
	\caption{Loss Functions: CE, CBCE, and the proposed JAC loss are compared using the same CNN sub-network, PNet-MSA. The Dice similarity coefficient (DSC) and jaccard index (JI) are reported in the form of mean $\pm$ standard deviation [worst case, best case].}%
	\label{tab:compare_losses}
}
\end{floatrow}
\end{figure*}
We then propose a new Jaccard (JAC) loss to train the neural network based image segmentation model. To optimize the Jaccard index (JI, a main segmentation metric) directly in network training, it makes the learning and inference procedures consistent and generate threshold-free segmentation. JAC loss is defined as follows:
\begin{eqnarray}
\ell_{jac} & {}={} & \mathcal{L}(\hat{Y}, Y) \nonumber \\
& {}={} & 1 - \frac{|Y_+ \bigcap \hat{Y}_+|}{|Y_+ \bigcup \hat{Y}_+|} = 1 - \frac{\sum_{j \in Y} y_j \land \hat{y}_j}{\sum_{j \in Y} y_j \lor \hat{y}_j} \\
& {}={} & 1 - \frac{\sum_{f \in Y_+} (1 \land \hat{y}_f)}{|Y_+| + \sum_{b \in Y_-} (0 \lor \hat{y}_b)},
\end{eqnarray}
where $Y$ and $\hat{Y}$ represent the ground truth and PNet-MSA prediction, which can be seamlessly replaced by the BiRNN's output $\bar{Y}$ and we will mainly focus on $\hat{Y}$ in the following discussion. Respectively, we have $Y_+$ and $Y_-$ defined as the foreground background pixel set, and $|Y_+|$ is the cardinality of $Y_+$. Similar definitions are also applied to $\hat{Y}$. $y_j$ and $\hat{y}_j \in \{0,1\}$ are indexed pixel values in $Y$ and $\hat{Y}$. In practice, $\hat{y}_j$ is relaxed to the probability number in the range of $[0,1]$ so that JAC loss can be approximated by
\begin{eqnarray}
\tilde{\ell}_{jac} & {}={} & 1 - \frac{\sum_{f \in Y_+} \min(1,\hat{y}_f)}{|Y_+| + \sum_{b \in Y_-} \max(0,\hat{y}_b)} \\
& {}={} & 1 - \frac{\sum_{f \in Y_+}\hat{y}_f}{|Y_+| + \sum_{b \in Y_-}\hat{y}_b}.
\end{eqnarray}
Theoretically, $L_{jac}$ and $\tilde{L}_{jac}$ share the same optimal solution of $\hat{Y}$. With slight abuse of notation, we use $L_{jac}$ to denote both. The model is then updated by gradient flows as,
\begin{equation}
\frac{\partial \ell_{jac}}{\partial \hat{y}_j} = \left\{ \begin{array}{rcl}
-\frac{1}{|Y_+|+\sum_{b\in Y_-} \hat{y}_b}, &~\mbox{for}~& j \in Y_+ \\
\frac{\sum_{f\in Y_+}\hat{y}_f}{(|Y_+| + \sum_{b\in Y_-}\hat{y}_b)^2}, &~\mbox{for}~& j \in Y_-
\end{array}\right.
\end{equation}
Since the inequality $\sum_{f\in Y_+}\hat{y}_f < (|Y_+|+\sum_{b\in Y_-} \hat{y}_b)$ holds by definition, the JACLoss assigns larger gradients to foreground pixels that intrinsically balances the foreground and background classes. It empirically works better than the cross-entropy loss or the classed balanced cross-entropy loss \cite{xie2015holistically} when segmenting small objects, such as pancreas in CT/MRI images. Similar loss functions are independently proposed and utilized in~\cite{MilletariNA16,ZhouXSFY16}. 

\section{Experimental Results and Analysis} \label{section:experiments}
\subsection{Experiment Set Up}
{\bf Datasets:} Two annotated pancreas datasets are utilized for experiments. The first one is the NIH-CT dataset~\cite{roth2015deeporgan,RothLFSS16} that is publicly available and contains 82 abdominal contrast-enhanced 3D CT scans. We organize a second MRI dataset~\cite{Cai2016}, with 79 abdominal T1-weighted MRI scans acquired under multiple controlled-breath protocol. For comparison, \emph{4-fold cross validation (CV)} is conducted similar to~\cite{roth2015deeporgan,RothLFSS16,Cai2016}. We measure the quantitative segmentation results using dice similarity coefficient (DSC):  $DSC=2(|Y_+\cap\hat{Y}_+|)/(|Y_+|+|\hat{Y}_+|)$, and Jaccard index (JI):  $JI=(|Y_+\cap\hat{Y}_+|)/(|Y_+\cup\hat{Y}_+|)$ as well as pixels-wise precision and recall. In addition, we use averaged Hausdorff distance (AVD) to evaluate the result of modeling inter-slice shape continuity and regularization.

{\bf Network Implementation:} Several variations of the fully convolutional network (FCN)~\cite{long2014fully} are evaluated. The first one, \ie{}, PNet, is proposed in our preliminary work~\cite{Cai2017Improving}, which is implemented as the holistically-nested network(HNN)~\cite{xie2015holistically}, but in a much smaller model size. Then, we have the improved PNet-MSA trained from scratch following the descriptions in Sec.~\ref{section:method}. For comparison, the holistically nested network (HNN)~\cite{xie2015holistically}, which is originally proposed for semantic edge detection is adapted for pancreas segmentation and been proved with good performance in medical object segmentations~\cite{RothLFSS16,Harrison_2017}. We also implement the UNet~\cite{Ronneberger2015} which is designed for medical image segmentation problems, where we find transferring UNet pre-trained model from other medical tasks is not as good as we train a new PNet-MSA from scratch for pancreas segmentation. Lower layers of HNN are transferred from VGG16 while UNet parameters are initiated from the snapshot released in~\cite{Ronneberger2015}. 

\begin{table*}[t!]
	\centering
	\caption{Comprison to different architectures: we have the proposed PNet-MSA compared with HNN network~\cite{xie2015holistically}, UNet~\cite{Ronneberger2015} and its preliminary version PNet-64~\cite{Cai2017Improving} under the same experimental setting up that the number of convoultional layer channel is 64. PNet-64 and PNet-MSA are trained from scratch, while HNN and U-Net are fine-tuned from released snapshots of other image domains.}
	\label{tab:compare_network_architecture}
		\begin{tabular*}{.9\textwidth}{l@{\extracolsep{\fill}}cccc} \hline

			\multirow{2}{*}{\bf Method} 
			& \multicolumn{4}{c}{mean$\pm$stdv. [min,max]} \\ \cline{2-5}
			& DSC(\%)  & JI(\%) & Precision & Recall \\ \hline
			\multicolumn{5}{c}{\bf CT} \\ \hline
			HNN~\cite{xie2015holistically} 	  & 79.6$\pm$7.7 [41.9,88.0] & 66.7$\pm$9.40 [26.5,78.6] & 83.4$\pm$6.5 [62.0,94.9] & 77.4$\pm$11.6 [28.3,92.6]  \\
			UNet~\cite{Ronneberger2015}       & 79.7$\pm$7.6 [43.4,89.3] & 66.8$\pm$9.60 [27.7,80.7] & 81.3$\pm$7.5 [49.6,97.0] & 79.2$\pm$11.3 [38.6,94.1]  \\
			PNet-64~\cite{Cai2017Improving}      & 81.3$\pm$8.1 [42.9,89.5] & 69.1$\pm$10.0 [27.3,81.0] & 82.8$\pm$6.5 [51.0,95.2] & 80.8$\pm$11.3 [28.5,93.8]  \\
			PNet-MSA        & \textBF{83.3$\pm$5.6 [59.0,91.0]} & \textBF{71.8$\pm$7.70 [41.8,83.5]} & \textBF{84.5$\pm$6.2 [60.7,96.7]} & \textBF{82.8$\pm$8.37 [56.4,94.6]}  \\ \hline
			\multicolumn{5}{c}{\bf MRI} \\ \hline
			HNN~\cite{xie2015holistically}	  & 75.9$\pm$10.1 [33.0,86.8] & 62.1$\pm$11.3 [19.8,76.6] & \textBF{84.4$\pm$6.4 [61.0,93.5]} & 70.6$\pm$13.3 [20.7,88.2] \\
			UNet~\cite{Ronneberger2015}    & 79.9$\pm$7.30 [54.8,90.5] & 67.1$\pm$9.50 [37.7,82.6] & 83.7$\pm$6.9 [64.6,94.6] & 77.3$\pm$10.3 [46.1,94.8] \\
			PNet-64~\cite{Cai2017Improving}    & 76.3$\pm$12.9 [6.30,88.8] & 63.1$\pm$14.0 [3.30,79.9] & 80.0$\pm$9.6 [41.6,92.5] & 74.6$\pm$15.6 [3.40,92.4] \\
			PNet-MSA		   & \textBF{80.7$\pm$7.40 [48.8,90.5]} & \textBF{68.2$\pm$9.64 [32.3,82.7]} & 84.3$\pm$7.6 [55.8,95.8] & \textBF{78.3$\pm$10.2 [38.6,95.0]} \\ \hline
		\end{tabular*}
\end{table*}

~~~Hyper-parameters are determined via model selection with training dataset. The PNet-MSA architecture containing 5 unit modules with 64 output channels in each convolution/deconvolution layer produces the best empirical performance while remaining with the compact model size ($<$3 million parameters, see Fig.~\ref{fig:pnet-architecture} for visual depicts). In our preliminary work~\cite{Cai2017Improving}, we also expand the number of output channels of convolutional/deconvolution layers to 128, however, it brings marginal improvement of PNet-MSA while dramatically increases computational load, thus, in this work, we fixed the number of output channels to 64 for all experiments. The training dataset is first split into a training-subset for network parameter training and a validation-subset for hyper-parameter selection. Denote the training accuracy as $Acc_{t}$ after model selection, we then combine then training- and validation-subset together to further fine-tune the network until its performance on the validation subset converges to $Acc_{t}$. All of our deep learning implementations are built upon Tensorflow~\cite{tensorflow2015-whitepaper} and Tensorpack~\cite{wu2016tensorpack}. The averaged time of model training is approximately 5 hours that operated on a single standard `` GeForce GTX TITAN'' GPU.

\subsection{Training Loss Comparison} \label{exp:loss}
Table~\ref{tab:compare_losses} presents results of three losses, \ie{}, the cross-entropy (CE), the class-balanced cross entropy (CBCE)\cite{xie2015holistically}, and the proposed Jaccard loss (JAC), under 4-fold cross-validation with the same PNet-MSA segmentation model. On the CT dataset, JAC outperforms CE and CBCE by 0.5\% and 0.2\% mean DSC, respectively. On the MRI dataset, JAC also achieves the best performance referring to the mean DSC and JI. We also evaluate the stability of segmentation performance with various thresholds. That is because the CNN network outputs probabilistic image segmentation maps instead of binary masks and an appropriate probability threshold is required to obtain the final binarized segmentation outcomes. However, it is often non-trivial to find the optimal probability threshold in practice. Empirically, Na\"{i}ve cross-entropy loss assigns the same penalty on positive and negative pixels so that the probability threshold should be around 0.5. Meanwhile, CBCE gives higher penalty scores on positive pixels (due to its scarcity), making the resulted ``optimal'' threshold at a relatively higher value. By contrast, JAC pushes the foreground pixels to the probability of 1.0 while remains to be strongly discriminative against the background pixels. Thus, JAC's plateau around the optimal segmentation performance would be much wider than CE and CBCE so that it could perform stably in a wide range of thresholds, \ie{}, $[0.05,0.95]$ in our experiments. Fig.~\ref{fig:losscurves} visually depicts the results of our analysis that the probability output maps from the JAC loss delivers the steadiest segmentation results referring to different output thresholds. 

\subsection{CNN Sub-Network Comparison}
Before modeling the shape continuity and regularization among image slices, it is important to first train an effective CNN sub-network, which presents the base for shape learning. We test multiple state-of-the-art CNN architectures, \ie{}, HNN~\cite{xie2015holistically}, UNet~\cite{Ronneberger2015}, and PNet~\cite{Cai2017Improving} with the proposed PNet-MSA. Each architecture design would have its own pros and cons and would perform differently according to the specific applications. It is possible that when fine-tuned with large-scale training data, the ImageNet pre-trained large-size networks, \eg{}, HNN and UNet, might outperform our compact-sized PNet and PNet-MSA. However, in this paper, we would focus on the situation that only limited amount of training data is available, which often happens in clinical practice. 

\begin{figure}[t!]
		\centering
		\subfloat[CT: U-Net]{%
			\includegraphics[width=.45\linewidth]{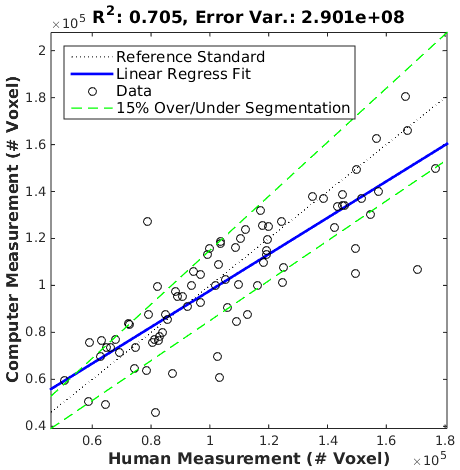}
		}
		\subfloat[CT: PNet-MSA]{%
		\includegraphics[width=.45\linewidth]{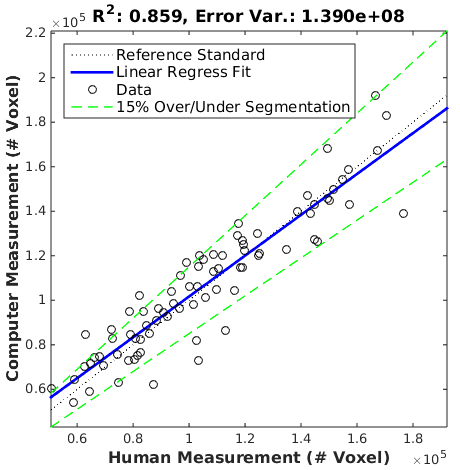}
		}	
		\\
		\subfloat[MRI: U-Net]{%
			\includegraphics[width=.45\linewidth]{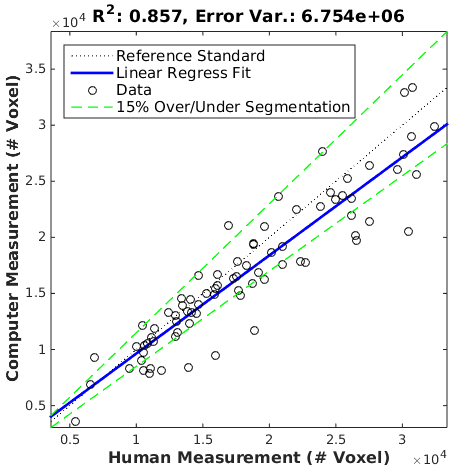}
		}
		\subfloat[MRI: PNet-MSA]{%
			\includegraphics[width=.45\linewidth]{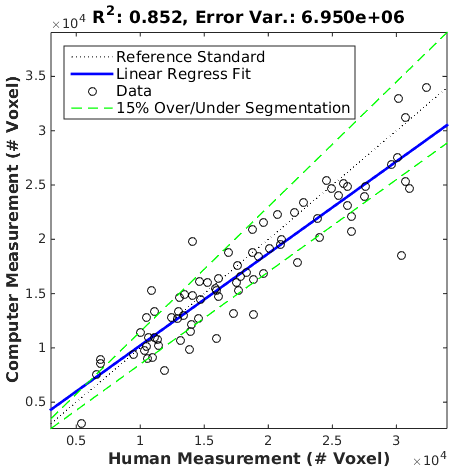}
		}
	\caption{Volume measurement: CT and MRI automatic measurements are presented in rows one and two, respectively. $R^2$ is the coefficient of determination of linear regression model, a higher $R^2$ value indicates better alignment between manual and automated measurement. Additionally, the variances of computer measurement error (Error Var.) are measured for UNet \cite{roth2015deeporgan}, and the proposed PNet-MSA. In each plot chart, the circles ($\circ$) represent data points and the solid line (in blue color) is the fitted line resulted from data regression. The reference standard and the 15\% over-/under-segmentation are presented in the datashed black and green lines. Better viewed in color with zoom.}
	\label{fig:volume-plot}
\end{figure}

~~~Table~\ref{tab:compare_network_architecture} presents segmentation results of different architectures. Without loss of generality, we set the output threshold for all CNN outputs to 0.5 basing on the analysis in Sec.~\ref{exp:loss}. In comparison, the proposed PNet achieves the best performance on both of the CT and MRI datasets. We then confirm the pancreatic volumes that measured by deep learning models is consistent with manual measurements in Fig.~\ref{fig:volume-plot}. PNet-MSA model trained from scratch output results demonstrate the best alignment between human and computerized measurements.  

\subsection{Modeling Shape Continuity Regularization} 
Given outputs of PNet-MSA as the best CNN-based segmentation results, the bi-direction RNN (BiRNN) sub-network is then stacked to the top of PNet-MSA and trained end-to-end. In each direction, a one-layer CLSTM is implemented with 1 hidden state and $3\times 3$ convolution filter kernels~\cite{ShiCWYWW15}. Particularly, the number of hidden state is set to 1 since our shape continuity learning is inherently simplified by processing only the output probability maps of CNN sub-network. CNN output $\hat{Y}_{\tau} \in R^{d^1_1\times d^2_1\times 1}$ (where $d^1_1$ and $d^2_1$ are the width and height of the input image) provides a highly compacted representation of the input CT/MRI image for shape learning. Thus, BiRNN with the hidden state $H_{\tau} \in R^{d^1_1\times d^2_1\times 1}$ is sufficient to model and capture the shape continuity regularization among CNN outputs. We notice that BiRNN could not converge stably during model training when a larger hidden state is used. In addition, we attempt to employ BiRNN on the feature maps from CNN's intermediate layers. Subsequently, this causes the model training process often not to converge. Thus, we mainly focus on the current design of BiRNN, which is simplified to learn the inter-slice shape continuity among successive CNN's segmentation outputs.

~~~In this work, we model the segmentation shape continuity as a higher-order inter-slice regularization among the CT/MRI axial slices. The average physical slice thickness in CT and MRI are 2mm and 7mm, respectively. Thus, slight shape change occurs between two correct segmented slices. Given the measured averaged Hausdorff distance (AVD) of neighboring slices in ground truth, the mean$\pm$standard deviation of shape changes in CT and MRI  are $0.35{}\pm{}0.94 $mm and $2.69{}\pm{}4.45$mm, respectively. The drastic shape changes in MRI volumes indicates that successive MRI image slices are actually more independent, so that in our implementation, shape continuity learning brings marginal but consistent performance improvement. The performance improvement in CT images is more evident. More specifically, we detect abnormal shape changes in the outputs of CNN and have them refined by BiRNN. We define abnormal shape change occurs between two neighboring CT when $AVD(\hat{Y}_{\tau},\hat{Y}_{\tau-1}){}>{}0.5$mm, which is decided basing on the shape change statics in the CT dataset. 

\begin{table}[t!]
	\centering
	\caption{Evaluate pancras segmentation on the CT dataset. BiRNN refines output of PNet-MSA providing better performance in both volume measurement (DSC) and surface reconstruction (AVD).}
	\label{tab:compare_birnn}
	\begin{tabular*}{\textwidth}{lcc} \hline
		\multirow{2}{*}{\bf Method} 
		& \multicolumn{2}{c}{mean$\pm$stdv. [min,max]} \\ \cline{2-3}
		& AVD (mm) & DSC (\%) \\ \hline 
		PNet-MSA & 0.61$\pm$0.53[0.15,3.48] & 83.3$\pm$5.6[59.0,91.0] \\
		BiRNN & \textBF{0.54$\pm$0.53[0.12,3.78]} & \textBF{83.7$\pm$5.1[59.0,91.0]}  \\ 
		\hline
	\end{tabular*}
\end{table}

\begin{figure}[t!]
	\centering
	\subfloat{\includegraphics[width=\linewidth]{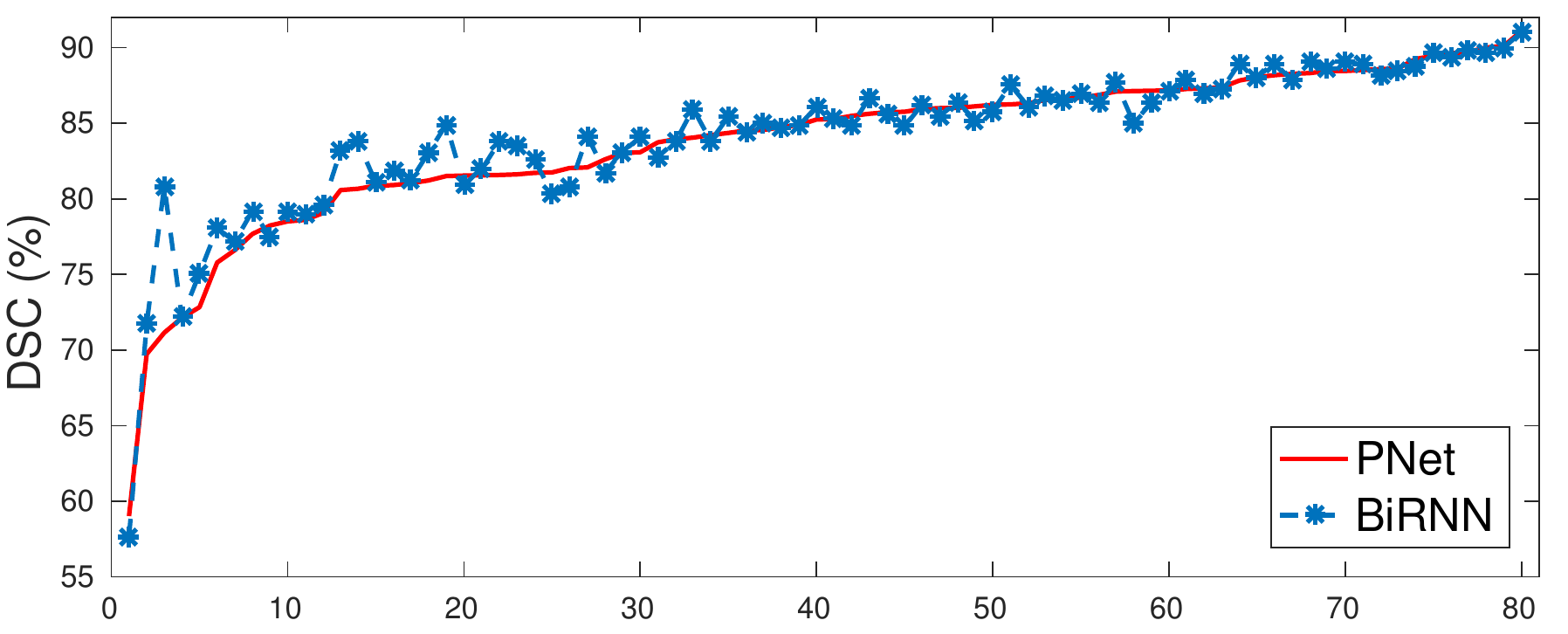}}
	\caption{80 cases with/without BiRNN refinement are sorted left to right by DSC scores of PNet-MSA, which is trained without explictly shape continuity constrains. Small fluctuations among the well segmented are possibly resulted from model updating, which can be ommitted as noise.}
	\label{fig:cl-improve}
\end{figure} 

~~~Table\ref{tab:compare_birnn} illustrates performance with and without shape continuity learning, where BiRNN boost volume segmentation (\ie{}, DSC) by 0.4\%. More importantly, the error for pancreatic surface reconstruction (\ie{}, AVD) drops from 0.61mm to 0.54mm, improved by 11.5\%. Fig.~\ref{fig:cl-improve} further shows the segmentation performance difference statistics, with or without contextual learning in subject-wise. In particular, those cases with low DSC scores are greatly improved by BiRNN. 

\begin{figure*}[t!]
	\centering
	\subfloat{%
		\includegraphics[width=.14\linewidth]{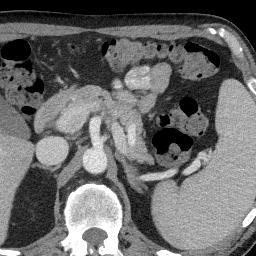}
	}
	\subfloat{%
		\includegraphics[width=.14\linewidth]{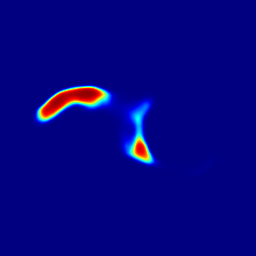}
	}
	\subfloat{%
		\includegraphics[width=.14\linewidth]{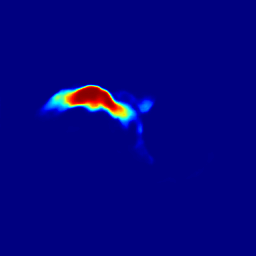}
	}
	\subfloat{%
		\includegraphics[width=.14\linewidth]{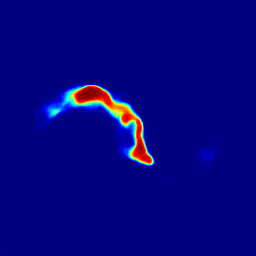}
	}
	\subfloat{%
		\includegraphics[width=.14\linewidth]{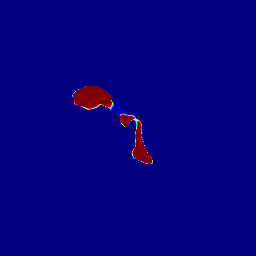}
	}
	\subfloat{%
		\includegraphics[width=.14\linewidth]{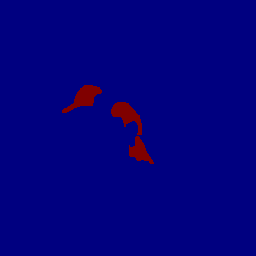}
	}
	\\[-2.5mm]
	\subfloat{%
		\includegraphics[width=.14\linewidth]{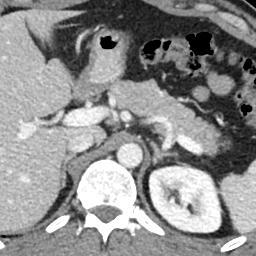}
	}
	\subfloat{%
		\includegraphics[width=.14\linewidth]{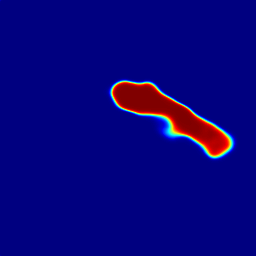}
	}
	\subfloat{%
		\includegraphics[width=.14\linewidth]{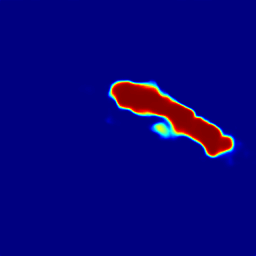}
	}
	\subfloat{%
		\includegraphics[width=.14\linewidth]{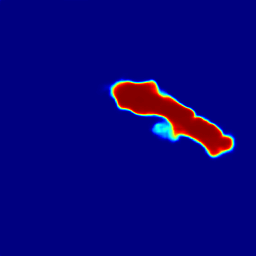}
	}
	\subfloat{%
		\includegraphics[width=.14\linewidth]{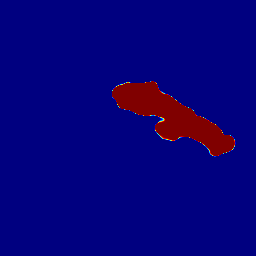}
	}
	\subfloat{%
		\includegraphics[width=.14\linewidth]{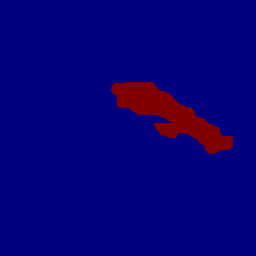}
	}
	\\[-2.5mm]
	\subfloat{%
		\includegraphics[width=.14\linewidth]{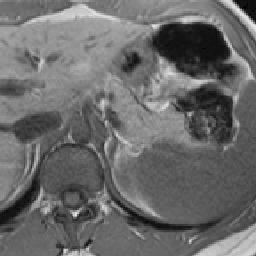}
	}
	\subfloat{%
		\includegraphics[width=.14\linewidth]{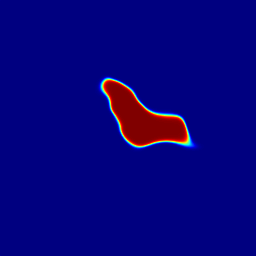}
	}
	\subfloat{%
		\includegraphics[width=.14\linewidth]{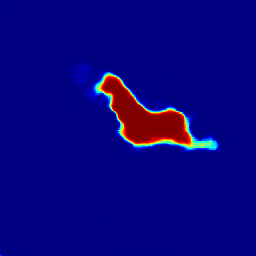}
	}
	\subfloat{%
		\includegraphics[width=.14\linewidth]{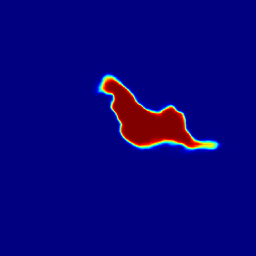}
	}
	\subfloat{%
		\includegraphics[width=.14\linewidth]{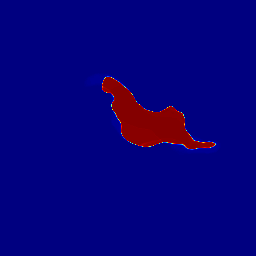}
	}
	\subfloat{%
		\includegraphics[width=.14\linewidth]{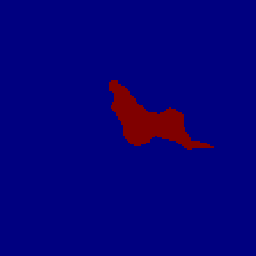}
	}
	\\[-2.5mm]
	\subfloat{%
		\includegraphics[width=.14\linewidth]{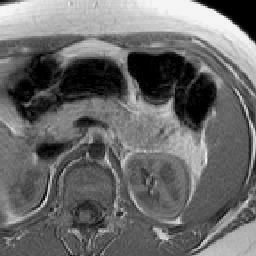}
	}
	\subfloat{%
		\includegraphics[width=.14\linewidth]{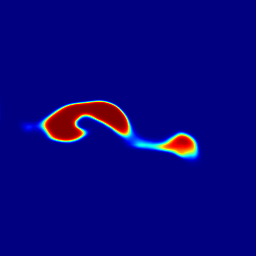}
	}
	\subfloat{%
		\includegraphics[width=.14\linewidth]{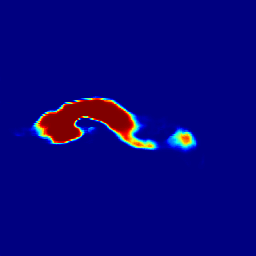}
	}
	\subfloat{%
		\includegraphics[width=.14\linewidth]{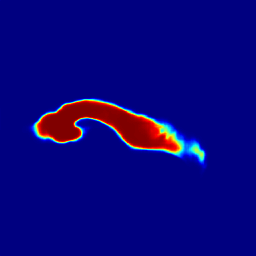}
	}
	\subfloat{%
		\includegraphics[width=.14\linewidth]{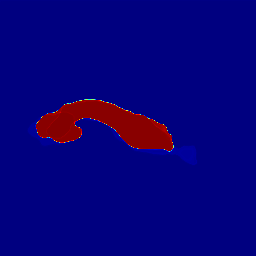}
	}
	\subfloat{%
		\includegraphics[width=.14\linewidth]{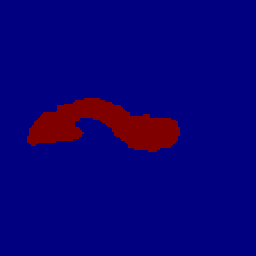}
	}
	\caption{Examples of output probability map: columns from left to right are the input CT/MRI image, results from HNN~\cite{xie2015holistically}, UNET~\cite{Ronneberger2015}, the proposed PNet-MSA sub-network, and the full CNN-RNN (``PNet-MSA+BiRNN''), and the ground truth. Our model delivers the most clear probability maps which preserve detailed pancreatic boundaries.}
	\label{fig:network-outputs}
\end{figure*}

~~~Finally, Fig.~\ref{fig:network-outputs} displays examples of output probability maps from all of the comparative methods, \ie{}, HNN~\cite{xie2015holistically}, UNet~\cite{Ronneberger2015}, PNet-MSA and ``PNet-MSA+BiRNN'', where the latter one delivers the sharpest and clearest output on both CT and MRI datasets. More specifically, PNet-MSA presents results that are detailed and recover the major part of the pancreas, where both HNN and UNet suffer from significant low segmentation recall. When observing the BiRNN outputs for CT and MRI, we find detailed pancreas parts in CT have been recovered via shape continuity learning and regularization, while in MRI, the BiRNN only outputs probability map with the same shape in PNet-MSA's output, which is optimal when the inter-slice shape changes drastically in the MRI dataset. Thus, BiRNN would help to refine pancreas segmentation with a smoothed surface in the situation that slice thickness of the 3D scans is reasonably small, \eg{}, $< 2$mm. 

\begin{table*}[t!]
	\centering
	\caption{Comparison with the state-of-the-arts: under 4-fold CV. PNet-MSA and BiRNN represent networks trained with JAC loss and constrains for inter-slice shape continuity, respectively. We show DSC and JI as mean $\pm$ standard dev. [worst, best]. The best result on CT and MRI are reported in respective by the proposed full CNN-RNN architecture, PNet-MSA+BiRNN and the PNet-MSA network with bold font.}
	\label{tab:cpr-state-of-the-art}
	\begin{tabular}{lcc} \hline
		\multirow{2}{*}{\bf Method} 
		& \multicolumn{2}{c}{mean$\pm$std. [min, max]} \\ \cline{2-3}
		& DSC(\%) & JI(\%) \\ \hline
		\multicolumn{3}{c}{\bf CT} \\ \hline
		3D FCN~\cite{2018arXiv180305431R}	  & 76.8$\pm$9.4 [43.7,89.4] & - \\
		Roth \etal{},~\cite{RothLFSS16}     		   & 78.0$\pm$8.2 [34.1,88.6] & - \\
		Roth \etal{},~\cite{Roth2017}           		 & 81.3$\pm$6.3 [50.6,88.9] & 68.8$\pm$8.12 [33.9,80.1] \\
		Coarse-to-Fine~\cite{ZhouXSFY16}   	& 82.3$\pm$5.6 [62.4,90.8] & - \\
		PNet + RNN~\cite{Cai2017Improving}   & 82.4$\pm$6.7 [60.0,90.1] & 70.6$\pm$9.00 [42.9,81.9] \\
		PNet-MSA                         	   					   & 83.3$\pm$5.6 [59.0,91.0] & 71.8$\pm$7.70 [41.8,83.5] \\
		PNet-MSA + Bi-RNN 							  & \textBF{83.7$\pm$5.1 [59.0,91.0]} & \textBF{72.3$\pm$7.04 [41.8,83.5]} \\
		\hline
        \multicolumn{3}{c}{\bf UFL-MRI-79} \\ \hline
		Graph-Fusion~\cite{Cai2016} & 76.1$\pm$8.70 [47.4,87.1] & -                 \\ 
		PNet-128~\cite{Cai2017Improving} & 80.5$\pm$6.70 [59.1,89.4] & 67.9$\pm$8.90 [41.9,80.9] \\		
		PNet-MSA & \textBF{80.7$\pm$7.40 [48.8,90.5]} & \textBF{68.2$\pm$9.64 [32.3,82.7]}  \\
		\hline
	\end{tabular}
\end{table*} 

\subsection{Comparison with the State-of-the-art Methods}
\begin{figure}[!htb]
	\centering
	\subfloat[DSC: 60\%]{%
		\includegraphics[width=0.33\linewidth, height=1.4cm]{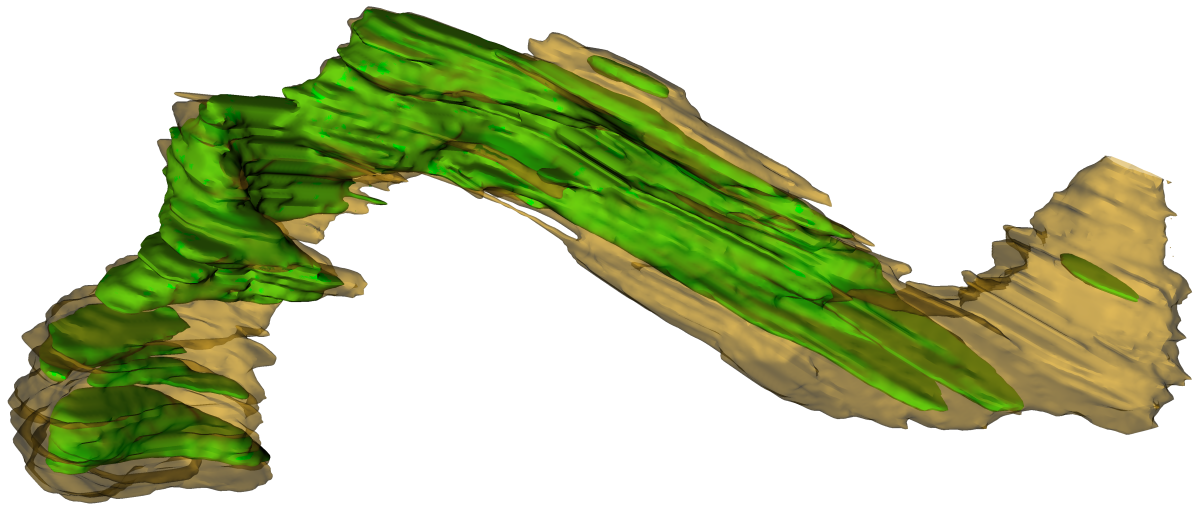}
	}
	\subfloat[DSC: 65\%]{%
		\includegraphics[width=0.33\linewidth, height=1.4cm]{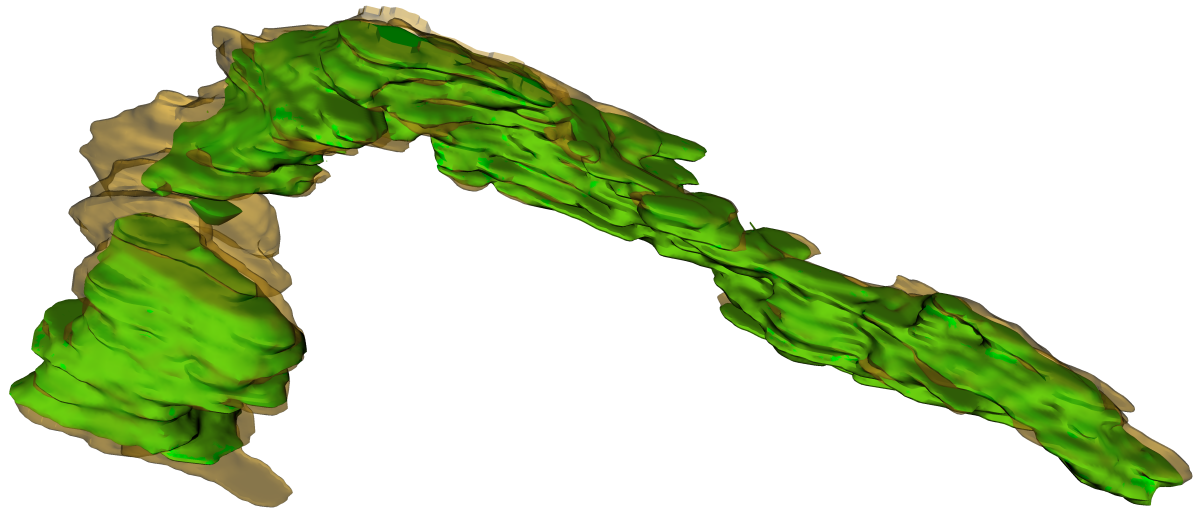}
	}
	\subfloat[DSC: 70\%]{%
		\includegraphics[width=0.33\linewidth, height=1.4cm]{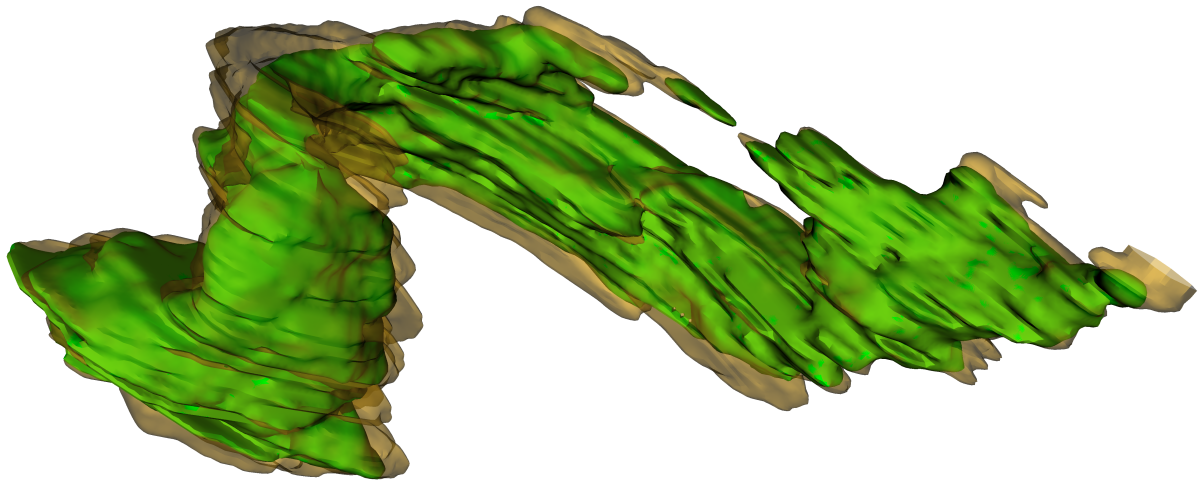}
	}
	\\[-2mm]
	\subfloat[DSC: 70\%]{%
		\includegraphics[width=0.33\linewidth, height=1.4cm]{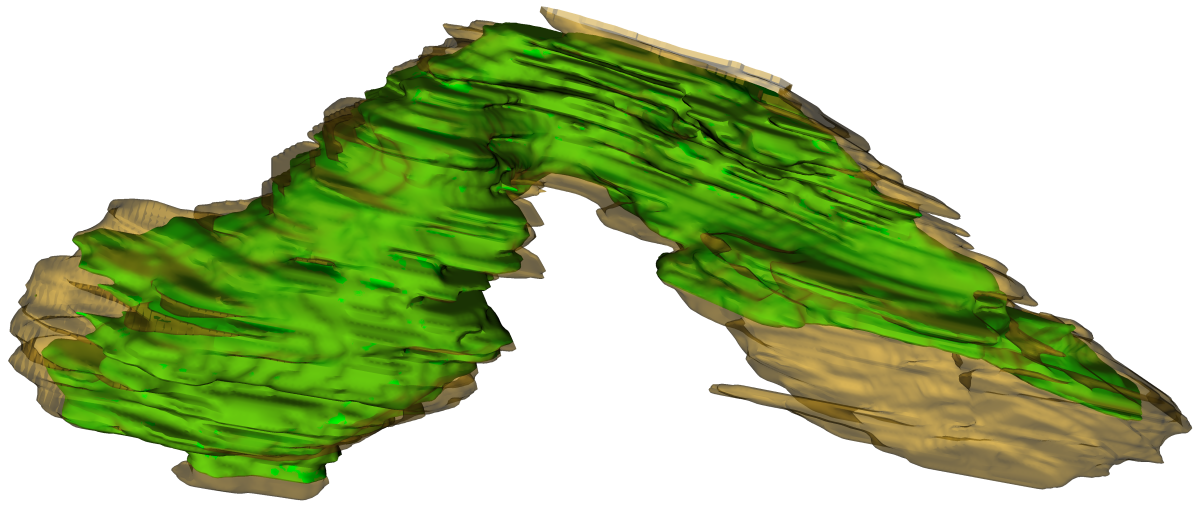}
	}
	\subfloat[DSC: 75\%]{%
		\includegraphics[width=0.33\linewidth, height=1.4cm]{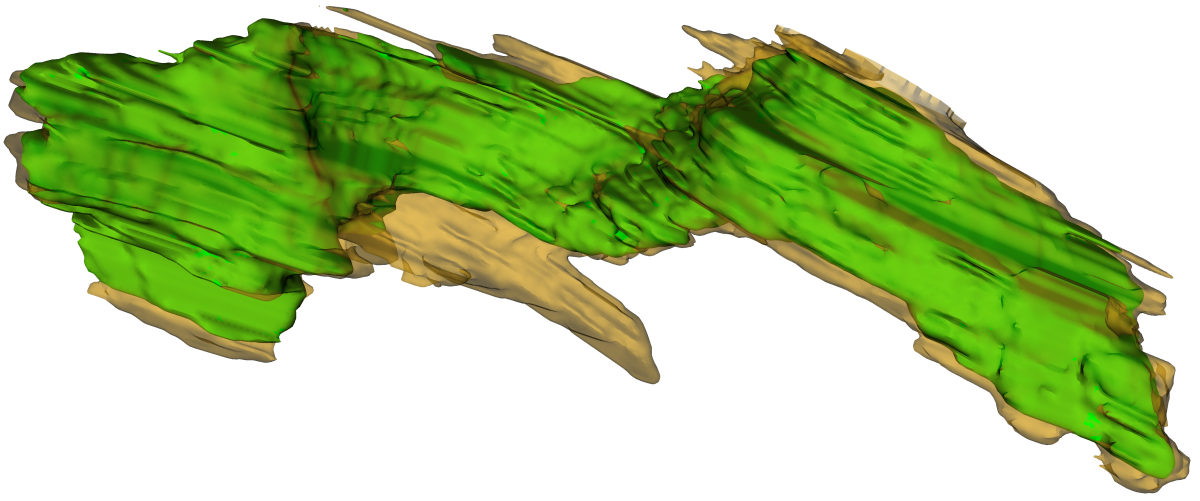}
	}
	\subfloat[DSC: 80\%]{%
		\includegraphics[width=0.33\linewidth, height=1.4cm]{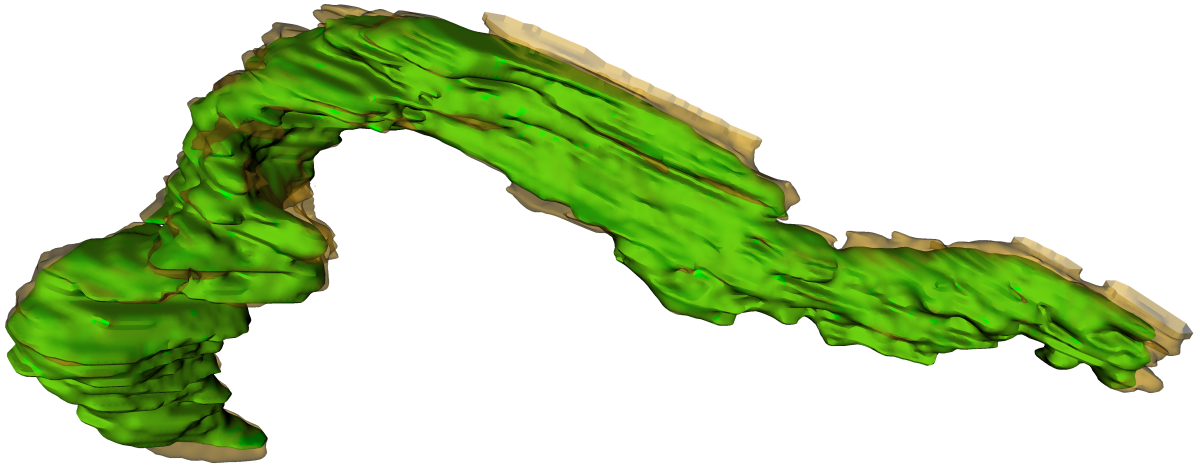}
	}
	\\[-2mm]
	\subfloat[DSC: 80\%]{%
		\includegraphics[width=0.33\linewidth, height=1.4cm]{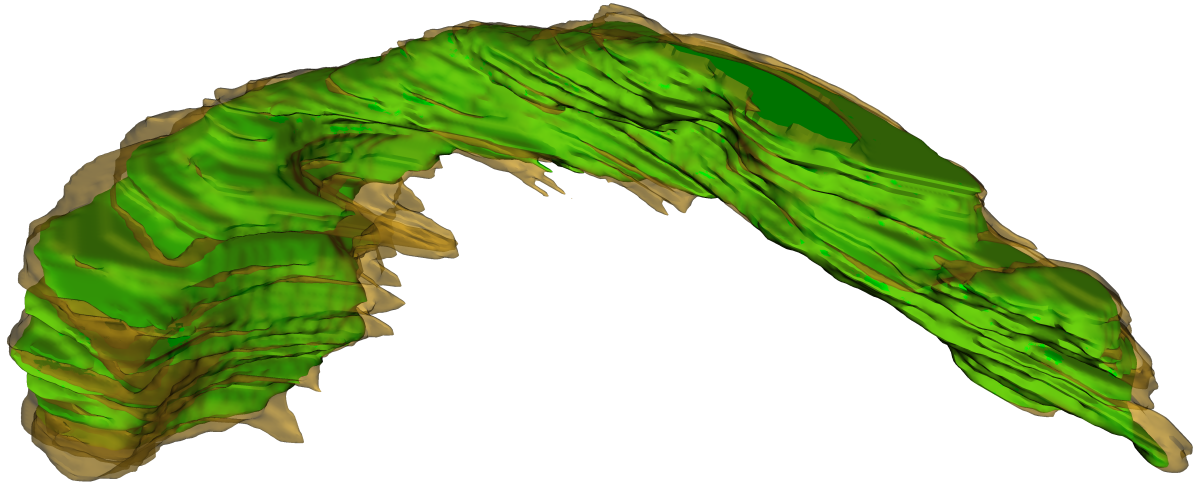}
	}
	\subfloat[DSC: 85\%]{%
		\includegraphics[width=0.33\linewidth, height=1.4cm]{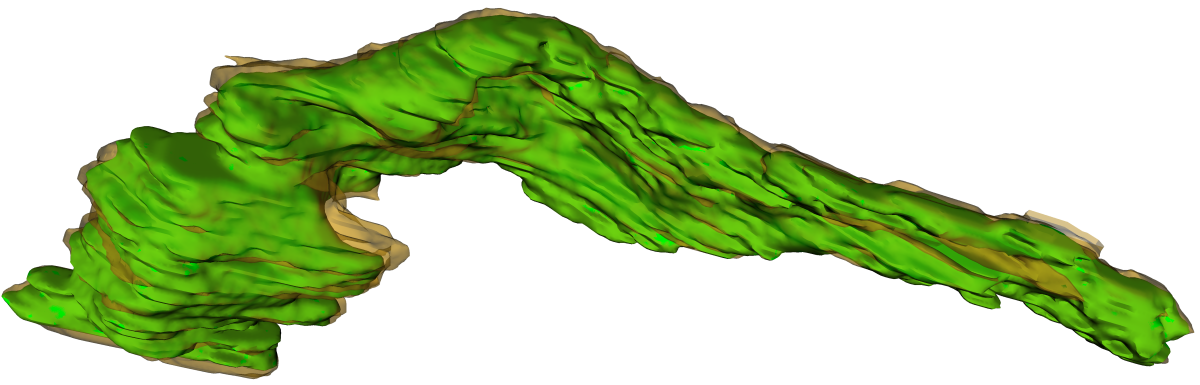}
	}
	\subfloat[DSC: 90\%]{%
		\includegraphics[width=0.33\linewidth, height=1.4cm]{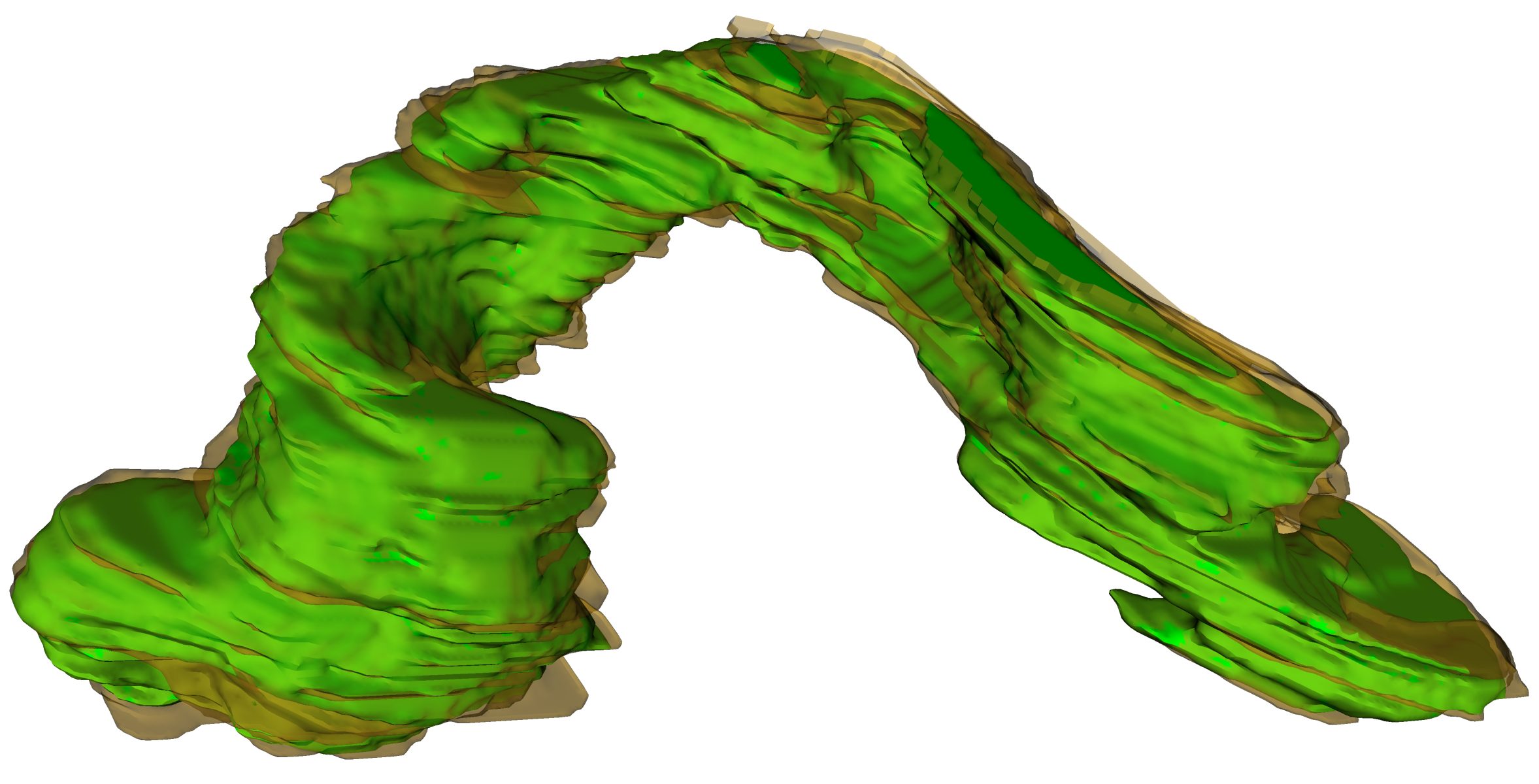}
	}
	
	\caption{3D visualization of pancreas segmentation results where human annotation shown in yellow and computerized segmentation displayed in green. The DSC are 90\%, 75\%, and 60\% for three examples from left to right, respectively.}
	\label{fig:3d-plot}
\end{figure}

We compare our pancreas segmentation models (as trained above) with the reported state-of-the-art. DSC and JI results computed from their segmentation outputs are reported in Table~\ref{tab:cpr-state-of-the-art}, under the same 4-fold CV. BiRNN performs the best on the CT dataset, and PNet-MSA achieves the best result on the MRI dataset. We notice that the current implementation of FCN 3D~\cite{2018arXiv180305431R} is not as effective as its 2D segmentation counterparts. The problem of segmenting 3D CT/MRI image volumes within a single inference is much more complex than the 2D CNN approaches where further network architecture exploration as well as more training images are typically required. This is referred as  ``curse of dimensionality'' in \cite{roth2015deeporgan,ZhouXSFY16}. In this scenario, we would argue that 2D network architectures may still be optimal for pancreas segmentation with large inter-slice thicknesses. We also note that our intuition of method development is orthogonal to the principles of ``coarse-to-fine'' pancreas location and detection~\cite{Roth2017,ZhouXSFY16}. Better performance may be achievable with the combination of both methodologies. Fig.~\ref{fig:3d-plot} visually depicts examples of reconstructed 3D segmentation results from the CT dataset. 

\section{Conclusion}
In this paper, we present a novel CNN-RNN architecture for pancreas segmentation in CT and MRI scans, via our tailor-made CNN module, followed by a bi-directional CLSTM to regularize the segmentation results on individual image slices. This is different from the independent process assumed in recent previous work~\cite{Cai2016,RothLFSS16,Roth2017,ZhouXSFY16}. The shape continuity regularization permits to enforce the pancreas segmentation spatial smoothness explicitly in the axial direction, in analogy to comprehending into videos by parsing and aggregating successive frames~\cite{Ng2015Beyond}. This may also share some similarity to the human doctor's way of reading radiology images. Combined with the proposed segmentation direct JAC loss function for CNN training to generate the threshold-free segmentation results, our quantitative pancreas segmentation results improve the previous state-of-the-art approaches \cite{Cai2016,RothLFSS16,Roth2017,ZhouXSFY16} on both CT and MRI datasets, with noticeable margins. Although our proposed method is focused on pancreas segmentation, this approach is generalizable to other organ segmentation in medical image analysis, in especially when the size of available training data is limited.



%

%



\ifCLASSOPTIONcaptionsoff
  \newpage
\fi



\bibliographystyle{IEEEtran}
\bibliography{IEEEabrv,./full,./ref}
\end{document}